\documentclass[runningheads]{llncs}
\usepackage[T1]{fontenc}
\usepackage{graphicx}
\usepackage{hyperref}

\usepackage[utf8]{inputenc}

\usepackage{microtype}

\usepackage{inconsolata}
\usepackage{graphicx}
\usepackage{multirow}
\usepackage{subfigure}
\usepackage{dsfont}
\begin{document}
\title{Exploring the Influence of Label Aggregation on Minority Voices: Implications for Dataset Bias and Model Training}
\titlerunning{Label Aggregation Bias in Datasets and Models}
%
\author{Mugdha Pandya\inst{1}\orcidID{0009-0003-5576-4921} \and
Nafise Sadat Moosavi\inst{1}\orcidID{0000-0002-8332-307X} \and
Diana Maynard\inst{1}\orcidID{0000-0002-1773-7020}}
\authorrunning{M. Pandya et al.}
%
\institute{Department of Computer Science, University of Sheffield, UK 
\email{\{mugdha.pandya, n.s.moosavi,d.maynard\}@sheffield.ac.uk}}
\maketitle              
\begin{abstract}
Resolving disagreement in manual annotation typically consists of removing unreliable annotators and using a label aggregation strategy such as majority vote or expert opinion to resolve disagreement. These may have the side-effect of silencing or under-representing minority but equally valid opinions. In this paper, we study the impact of standard label aggregation strategies on minority opinion representation in sexism detection. We investigate the quality and value of minority annotations, and then examine their effect on the class distributions in gold labels, as well as how this affects the behaviour of models trained on the resulting datasets. Finally, we discuss the potential biases introduced by each method and how they can be amplified by the models.
\end{abstract}
\section{Introduction}
Disagreements among annotators are common for any annotation task, especially for subjective tasks such as sexism detection. They arise for two main reasons: carelessness or misunderstanding of the task; and differing yet equally valid opinions (e.g. men and women may disagree on what constitutes a sexist comment) \cite{beck-etal-2020-representation}.

Although disagreement information has recently been incorporated into training \cite{plank2014learning} and evaluation \cite{davani2022dealing,gordon2022jury}, the predominant approach relies on label aggregation to generate gold labels \cite{zampieri-etal-2019-predicting}. 
Disagreements are typically resolved by taking the majority vote or relying on an expert \cite{davidson-etal-2021-improved,founta2018,waseem-hovy-2016-hateful}. However, these strategies can introduce bias into the dataset, as they tend to disregard differing annotations that represent the views of a minority of the annotators, stemming from different cultural and social contexts \cite{jiangplos,aroyo2013crowd}. 
It is potentially important to represent these minority voices in both data and downstream models, but there has been limited systematic investigation into how label aggregation strategies may introduce bias in this respect. The potential implications of this on models trained with such data also remain largely unexplored.

We address this gap by investigating different label aggregation strategies and their impact on minority opinion inclusion, as well as testing the downstream effects 
on model behaviour on 2 datasets for sexism detection. We also conduct a qualitative analysis to understand the type of disagreements between annotators. We define minority vote aggregation as choosing the labels that the fewest annotators chose for a post.\footnote{We make no statements about minority annotators or their protected attributes.}

Our results show that majority aggregation captures opinions well when sexism is clear, and there is little difference of opinion. However, this produces fewer occurrences of fine-grained but important classes such as dehumanisation and mistreatment of women, which may be important to recognise specifically due to their potentially more harmful nature \cite{posetti2021chilling}. This effect is amplified in the predictions of models that are trained on the aggregated data. Overall, our investigation highlights the need for consideration of the most suitable strategy according to the data and objective of the task.
\section{Related Work}
\label{sec:rw}
A common approach for dataset creation is to employ multiple annotators to label input and subsequently to derive a gold label using an aggregation strategy such as majority voting, averaging, or expert intervention \cite{sabou2014corpus,waseem-hovy-2016-hateful}. There are two types of annotations that deviate from the gold label: noisy labels and labels influenced by differing experiences, beliefs, and opinions, including minority viewpoints that may not align with the majority or expert opinions. Efforts have been made to identify and address noisy labels \cite{kazai2013,rottger2021two,sap2021annotators,beck-etal-2020-representation}.



Recently, studies have drawn attention to the under-representation of minority opinions \cite{prabhakaran2021releasing}, and researchers have criticised the notion of a single, definitive gold label \cite{mieleszczenko2023capturing,aroyo2015truth,plank2022problem,baan2022stop}, as these approaches tend to prioritise the majority viewpoint to achieve state-of-the-art (SOTA) performance scores. Nevertheless, SOTA systems often rely heavily on gold labels for achieving superior performance on standard evaluation metrics \cite{vitsakis-etal-2023-ilab,gordon2021}. In response, methods have been explored to enhance gold label generation. \cite{ren2023label} proposes label confidence-based noise corrections and incorporates new data, while \cite{wu2023learning} identifies labelling biases and trends (e.g., favouring a particular class) and implements adjustments. Moreover, \cite{wei2023aggregate} examines the downstream impact of aggregating and utilising multiple labels.

In this study, we investigate various label aggregation strategies employed to generate a single gold label, assessing their impact on minority annotations and analysing their downstream effects on model behaviour. 
\section{Datasets}
\label{sec:data}
We have selected two publicly available datasets, both providing individual annotations and gold labels obtained after resolving disagreements. Details of datasets in Table~\ref{tab:datasets}.
All sensitive information was anonymised or not made publicly available by the original authors of the datasets.

\begin{table}[t!]
\caption{Details of datasets}
\centering
%
\begin{tabular}{|l|l|l|}
\hline
\textbf{Dataset} & \textbf{EDOS} & \textbf{GE} \\\hline
\textbf{Lang} & English & English\\\hline
\textbf{Posts} & 20k & 13.6k\\\hline
\textbf{Platform} & Gab, Reddit & Twitter\\\hline
\textbf{Annotators} & 19 & 105 \\\hline
\textbf{Ann/post} & 3 & 5\\\hline
\textbf{Ann type} & expert & MTurk\\\hline
\textbf{Selection} & expert & convert \& maj\\\hline
\textbf{Number of labels}& T1-2, T2-4,T3-11 & 2\\\hline
\end{tabular}

\label{tab:datasets}
\end{table}

\subsection{Call me sexist, but...} 
This dataset (GE) \cite{Samory_Sen_Kohne_Flöck_Wagner_2021} combines multiple datasets to produce sexism classes. Each post is annotated for the sexist nature of its content and the incivility of the phrasing. The sexism label is derived by combining the two via a scoring system. Although the dataset contains 13.6k posts, we only use the 5.7k for which annotations from individual annotators are available.

\subsubsection{Binary label conversion:}
Each annotator annotates a post for content (C) and phrasing (P). If they choose classes C1 to C4 in C or P1 in P, their annotation for a post is sexist. If not, it is not sexist. A majority vote of the 5 annotators` $sexism$ label choices is used to generate the gold label.
The sexist nature of content classes C1 to C6 and the incivility of phrasing classes P1 to P3 from \cite{Samory_Sen_Kohne_Flöck_Wagner_2021} are as follows:
\begin{itemize}
    \item \textbf{Content C1-C6:} Behavioural expectations (C1); Stereotypes and comparative opinions(C2); Endorsement of inequality (C3); Denying inequality and rejecting feminism (C4); Maybe sexist: can't tell without context (C5); Not sexist: not a direct statement of sexist opinion (C6).
    \item \textbf{Phrasing P1-P3:} Uncivil and sexist - attacks, foul language, or derogatory depictions directed towards individuals because of their gender (P1); Uncivil but not sexist - offensive phrasing or message that is not clearly directed towards the target`s gender (P2); Civil (can be sexist or not sexist) - neutral phrasing that does not contain offenses or slurs (P3).
\end{itemize}




\subsection{Explainable Detection of Online Sexism}
\begin{figure}[hbt!]
 \centering
 \includegraphics[width=0.7\textwidth]{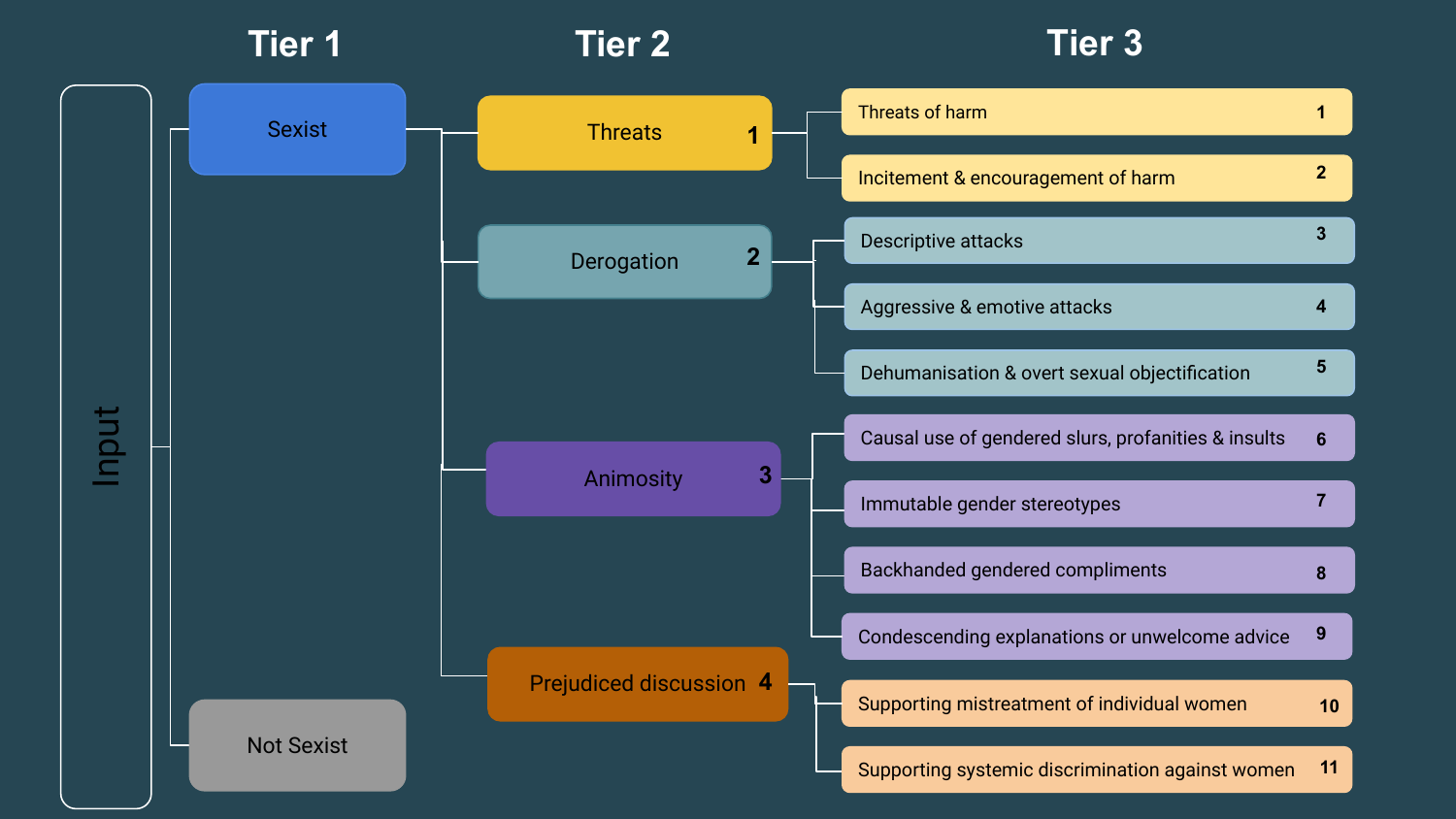}
 \caption{EDOS dataset label schema. Figure from \cite{kirk2023semeval}}
 \label{fig:edos_label}
\end{figure}

This dataset (EDOS) was created for the SemEval 2023 task 10 \cite{kirk2023semeval} on explainable detection of online sexism, featuring a hierarchical label schema as shown in Figure~\ref{fig:edos_label}. 
Tier 1 (T1) contains binary labels - sexist and not sexist. The posts labelled as sexist in T1 are classified into 4 labels in tier 2 (T2). These four labels are further classified into 11 labels in tier 3 (T3).

We focus on T3 results more than T1 and T2 since T3 contains fine-grained labels. By focusing on the fine-grained classes in T3, we can compare how the different aggregation techniques are able to capture nuanced sexism. Therefore, we discuss results for only T3 in the main body of the paper.Results for T1 and T2 are in Appendix~\ref{app:A}
\section{Investigating Aggregation on Data}
\label{sec:ss}
\begin{figure}
 \centering
 \subfigure{\includegraphics[width=0.58\linewidth]{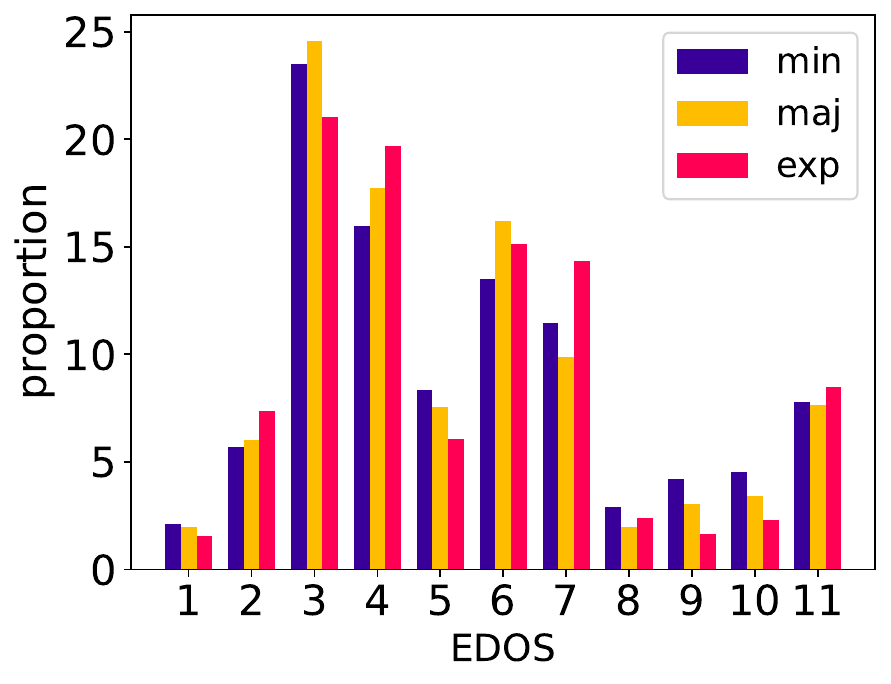}} 
 \subfigure{\includegraphics[width=0.4\linewidth,height=0.45\linewidth]{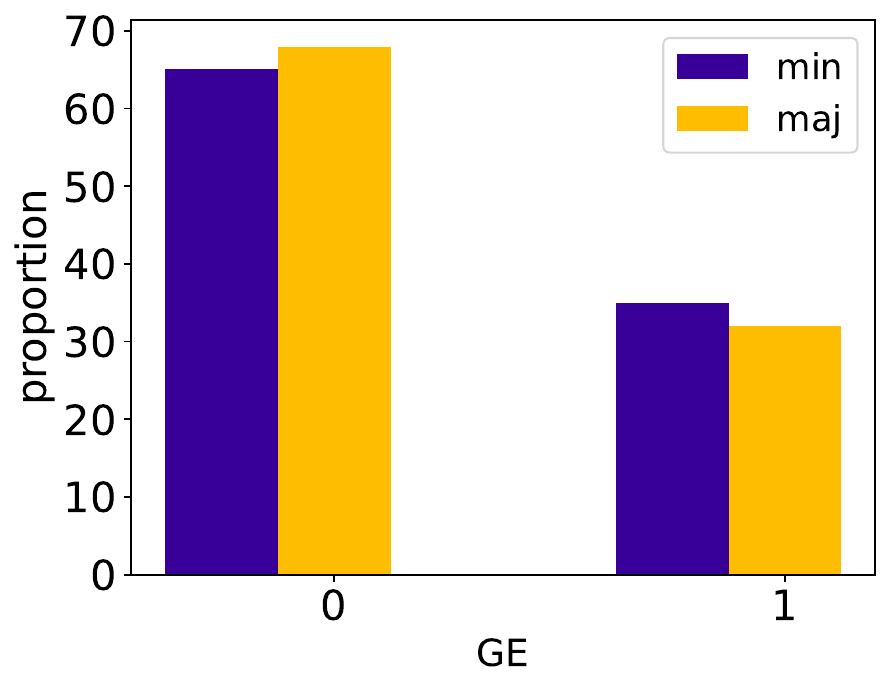}}
 \caption{Class distributions of data for EDOS and GE. EDOS class labels are threats of harm (1); incitement and encouragement of harm (2); descriptive attacks (3); aggressive and emotive attacks (4); dehumanisation and sexual objectification (5); casual use of gendered slurs, profanities and insults (6); immutable gender differences (7); backhanded gender compliments (8); condescending explanations or unwelcome advice (9); supporting mistreatment of individual women (10); and supporting systemic discrimination against women (11).}
 \label{fig:ss}
\end{figure}

We first investigate the impact of various label aggregation strategies on the distribution of labels within the dataset. In addition to majority vote and expert opinion\footnote{Only for EDOS - there is no expert annotation available for the GE dataset.}, we also explore minority aggregation, where the gold label is determined by selecting the minority vote (labels that the fewest annotators chose for a post). We select this method to understand the nature of bias and the role of the minority vote to explore how sensitive or easily overlooked interpretations might be missed by majority consensus. We present the distribution of different labels for the EDOS and GE datasets in Figure~\ref{fig:ss}.

We observe that \textbf{(a)} in GE, by using minority aggregation, more inputs are labelled as sexist, suggesting that some sexist posts may be overlooked with a majority vote; and \textbf{(b)} in EDOS, the ratio of certain labels, including threats of harm (1), dehumanisation and sexual objectification (5), backhanded gender compliments (8), condescending explanations or unwelcome advice (9), and supporting mistreatment of individual women (10), which are already not very frequent, would be further reduced in EDOS when using the majority or expert opinion. \footnote{Table~\ref{tab:ss_edos_model_app} and Figure~\ref{fig:edos_ss_model_app} for EDOS T1 \& T2 results in Appendix~\ref{app:A}}

Looking at these distributions, minority aggregation captures the more nuanced and harmful forms of sexism, while those chosen more frequently by the majority tend to be easy to identify. Examples can be found in Table~\ref{tab:nuance}. Expert aggregation also captures nuanced sexism more than majority aggregation but less than minority aggregation does. However, the risk of using this strategy is that the gold labels will then have the expert's biases. Another problem is that only a single label can be chosen in spite of multiple viable options, e.g., a post could be both dehumanising and contain gendered slurs. In this case, most annotators choose the more easily identifiable label.

\begin{table}[t!]
\caption{Examples of nuanced forms of sexism captured by minority aggregation}
\label{tab:nuance}
\footnotesize
\resizebox{\textwidth}{!}{
    \centering
    \begin{tabular}{|p{0.64\linewidth}|p{0.16\linewidth}|p{0.18\linewidth}|}
    \hline
    text & majority & minority\\\hline
    \textit{Jail Linda Cockroach Sarsour, press charges that stick and lock her up for a long time. Punish her personally. No letting her become a feminazi martyr}&6. Profanity&10. Mistreat\\\hline
    \textit{saw a lady driving full speed into a garage door so they are quite good at stunt work. Other than that they aren't very good at anything.}&2. Descriptive&8. Backhand comp\\\hline
    \textit{Ive definitely only see the two extremes, either super hot and not interested, or fat beasts, who make up like 90\%of hinge. that being said, my last two relationships were hinge girls and they were pretty attractive}&2. Descriptive&5. Dehumanise\\\hline
    \end{tabular}}
    
\end{table}

\subsection{Label Alignment between Aggregation Techniques}
We compare the ratio of the counts of the label generated via different aggregation techniques that align with the gold label (i.e. overlap with gold labels). For GE, the label generated via minority aggregation aligns with the gold label 66\% of the time (meaning all annotators agree 66\% of the time). Results for EDOS are in Table~\ref{tab:align}. We see that the majority vote is more similar to the expert in all three tiers. As expected, the minority vote differs greatly from the expert for fine-grained labels. 
\begin{table}[hbt!]
\caption{Label alignment between aggregation techniques for EDOS}
    \label{tab:align}
    \centering
    \begin{tabular}{|c|c|c|}
    \hline
    Tier&minority - expert&majority - expert\\\hline
    T1&85.09\%&96.64\%\\\hline
    T2&44.89\%&85.36\%\\\hline
    T3&31.26\%&84.37\%\\\hline
    \end{tabular}
    
\end{table}

\subsection{Comparing Nuanced Sexism Detection by Aggregation Technique}

We conduct an analysis confirming that minority labels often identify but less apparent perspectives. For the EDOS dataset, we find that in the minority aggregation, labels easily overlooked and not recognised by most annotators are better represented. In the minority aggregation, we find that the classes deemed more challenging to identify are chosen 26.13\% of the time. They are only chosen 10.88\%  and 12.02\% of the time when majority label aggregation and expert aggregation are used. Table~\ref{tab:min} shows the breakdown for each of the relevant labels. This analysis supports our initial hypothesis that minority labels,  in our evaluated datasets, often identify crucial but less apparent perspectives. These classes are also more harmful in nature \cite{posetti2021chilling} than easily identifiable labels, such as casual use of gendered slurs, profanities and insults (6).

\begin{table}[hbt!]
\caption{Breakdown of posts belonging to nuanced classes for each aggregation technique.}
    \label{tab:min}
    \centering
    \begin{tabular}
        {|l|c|c|c|}
        \hline
         Label&  minority (\%)&  majority (\%)& expert (\%)\\\hline
         1. Threats&  2.45&  0.92& 1.75\\\hline
         5. Dehumanise&  10.33&  6.54& 5.69\\\hline
         8. Backhand comp&  3.41&  0.71& 1.35\\\hline
         9. Condescend&  4.59&  0.92& 1.15\\\hline
         10. Mistreat&  5.36&  1.79& 2.08\\\hline
    \end{tabular}
\end{table}
\section{Investigating Aggregation on Models}
\label{sec:model}
To understand the effect of different label aggregation strategies on model performance, we fine-tuned BERT \cite{kenton2019bert}, and RoBERTa \cite{liu2019roberta} on each. We choose the model with the best validation loss.\footnote{Model parameters and performance metrics in Appendix~\ref{app:D}.}

Figure~\ref{fig:ss_model} shows the comparative class distributions for both models for GE and EDOS. Table~\ref{tab:ss} shows the class distributions of the test data for each strategy.
\begin{figure}[hbt!]
 \centering
 \subfigure{\includegraphics[width=0.58\linewidth]{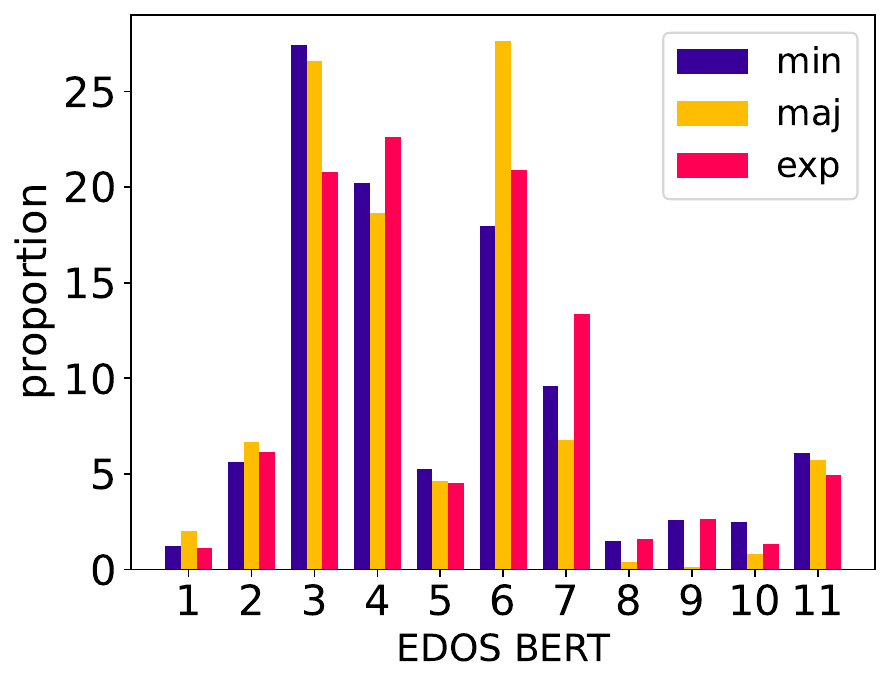}} 
 \subfigure{\includegraphics[width=0.4\linewidth,height=0.45\linewidth]{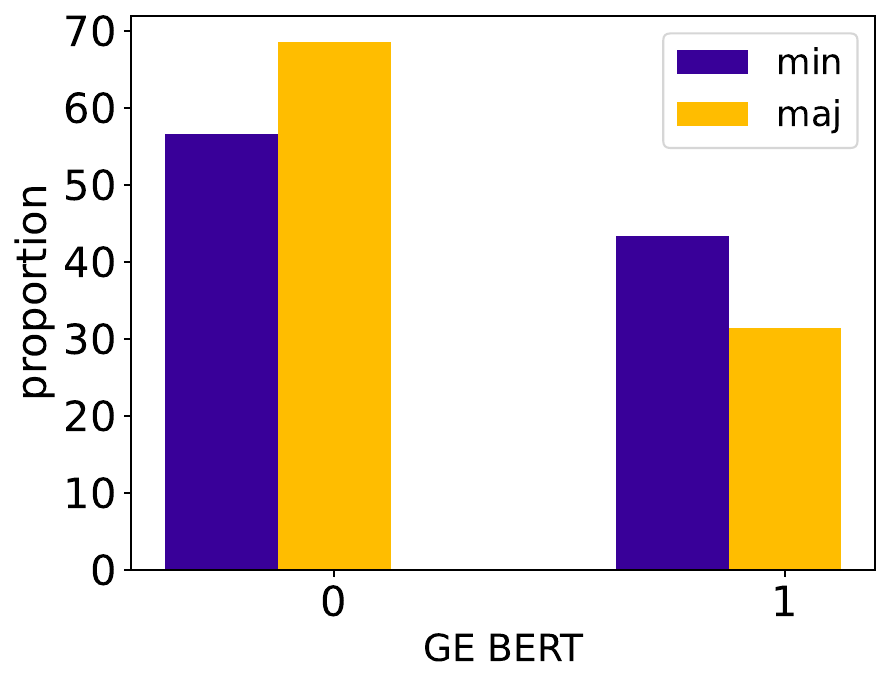}}
 \subfigure{\includegraphics[width=0.58\linewidth]{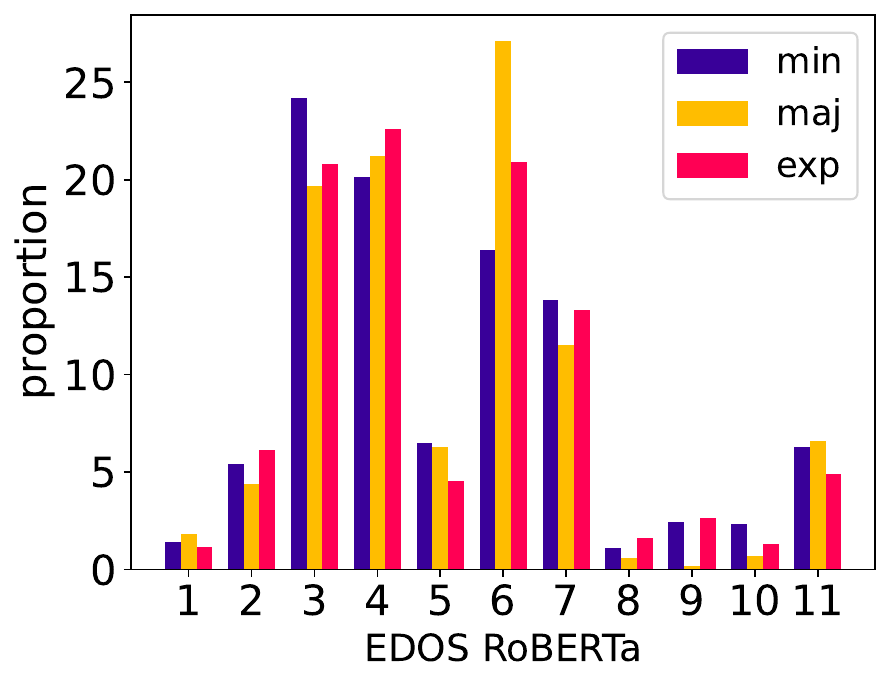}} 
 \subfigure{\includegraphics[width=0.4\linewidth,height=0.45\linewidth]{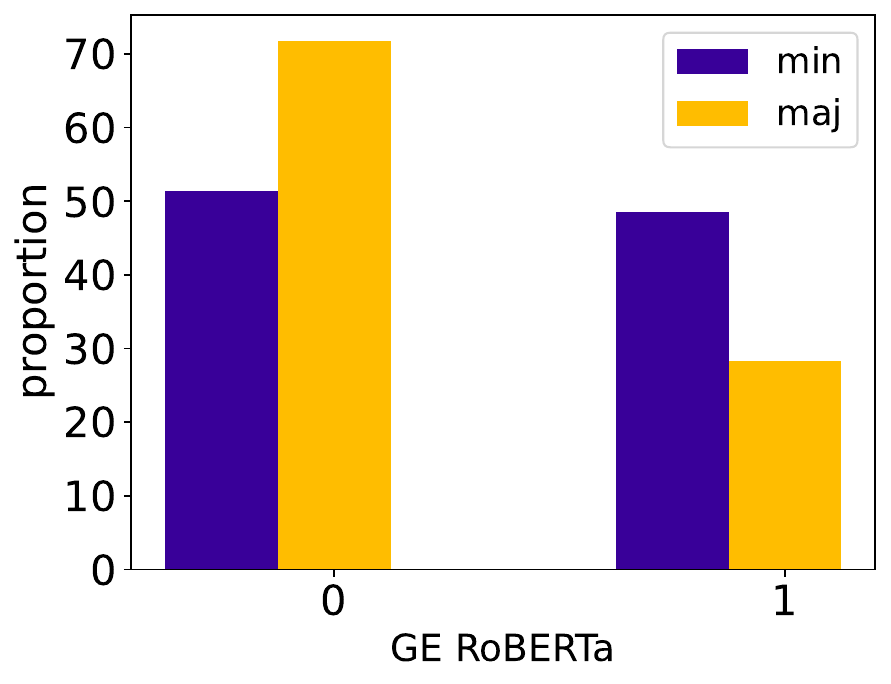}}
 \caption{Class distributions from model predictions}
 \label{fig:ss_model}
\end{figure}

We observe that \textbf{(a)} for both BERT and RoBERTa, there is a considerable difference between majority and minority aggregation for GE. More posts are identified as sexist when minority aggregation is used. Interestingly, the gap between the sexist and non-sexist classes also increases in the model prediction: for sexist, the difference between minority and majority in the data is 2.86\%, but in model prediction, it is between 9\% and 20\% points, indicating that these biases can be amplified by the model. \textbf{(b)} For both models, in EDOS, majority aggregation has a higher proportion of threats of harm (1), while it has very low counts for condescending explanations (9), backhanded compliments (8), and mistreatment of individual women (10). Minority aggregation has the highest proportion of descriptive attacks (2), dehumanisation and sexual objectification (5), condescending explanations (9), and supporting systemic discrimination of women (11). Unlike GE, the bias amplification is less obvious. This could be due to data sparsity for some labels, particularly those representing the minority, and so the model does not learn it.
\begin{table}
 \caption{Class distributions \% from test data \& model predictions \& difference from the original gold labels}
 \label{tab:ss}
 
\resizebox{\textwidth}{!}{
 \centering
 \begin{tabular}{|p{0.08\linewidth}|l|c|c|c|c|c|c|c|c|c|} \hline 
\multicolumn{1}{|c}{Dataset}& \multicolumn{1}{|c}{Class} &\multicolumn{3}{|c}{Data}& \multicolumn{3}{|c}{BERT} &\multicolumn{3}{|c|}{RoBERTa}\\ \hline
&& majority&minority&expert&majority&minority&expert&majority&minority&expert\\ \hline 
 \multirow{2}{0.08\linewidth}{\centering GE}&Not sexist&68.54&65.68$_{\downarrow2.86}$&-&68.54&56.63$_{\downarrow9.91}$&-&71.71&51.43$_{\downarrow20.28}$&-\\ \cline{2-11} 
 &Sexist&31.46&34.32$_{\uparrow2.86}$&-&31.46&43.37$_{\uparrow9.91}$&-&28.29&48.57$_{\uparrow20.28}$&-\\ \hline 
 \multirow{11}{0.08\linewidth}{\centering T3}&1. Threats&1.75&2.10$_{\uparrow0.35}$&1.75&1.98$_{\uparrow0.84}$&1.22$_{\uparrow0.04}$&1.14&1.82$_{\uparrow0.23}$&1.42$_{\downarrow0.17}$&1.59\\ \cline{2-11} 
 &2. Incitement&5.65$_{\downarrow1.10}$&5.53$_{\downarrow1.22}$&6.75&6.69$_{\uparrow0.54}$&5.62$_{\downarrow0.53}$&6.15&4.39$_{\downarrow3.00}$&5.39$_{\downarrow2.00}$&7.39\\ \cline {2-11}
 &3. Descriptive&25.37$_{\uparrow4.70}$&23.99$_{\uparrow3.32}$&20.67&26.60$_{\uparrow5.79}$&27.43$_{\uparrow6.62}$&20.81&19.68$_{\uparrow3.50}$&24.18$_{\uparrow8.00}$&16.18\\ \cline {2-11}
 & 4. Agg \& emo&18.64$_{\downarrow3.41}$&15.68$_{\downarrow6.37}$&22.05&18.66$_{\downarrow3.95}$&20.21$_{\downarrow2.40}$&22.61&21.20$_{\downarrow0.60}$&20.15$_{\downarrow1.65}$&21.80\\ \cline {2-11}
 &5. Dehumanise&8.14$_{\uparrow1.98}$&8.68$_{\uparrow2.52}$&6.16&4.63$_{\uparrow0.09}$&5.21$_{\uparrow0.67}$&4.54&6.28$_{\uparrow1.79}$&6.48$_{\uparrow1.99}$&4.49\\\cline {2-11}
 &6. Profanity&15.65$_{\uparrow1.04}$&13.57$_{\downarrow1.04}$&14.61&27.63$_{\uparrow6.72}$&17.98$_{\downarrow2.93}$&20.91&27.10$_{\uparrow4.74}$&16.39$_{\downarrow5.97}$&22.36\\ \cline {2-11}
 &7. Immutable diff&9.32$_{\downarrow3.91}$&11.52$_{\downarrow1.71}$&13.23&6.76$_{\downarrow6.67}$&9.60$_{\downarrow3.74}$&13.34&11.51$_{\downarrow4.11}$&13.83$_{\downarrow1.79}$&15.62\\ \cline {2-11}
 &8. Backhand comp&2.20$_{\downarrow0.93}$&3.43$_{\uparrow0.30}$&3.13&0.37$_{\downarrow1.24}$&1.49$_{\downarrow0.12}$&1.61&0.61$_{\downarrow1.64}$&1.09$_{\downarrow1.16}$&2.25\\ \cline {2-11}
 &9. Condescend&3.28$_{\uparrow1.10}$&4.11$_{\uparrow1.93}$&2.18&0.15$_{\downarrow1.32}$&2.60$_{\uparrow1.13}$&1.47&0.15$_{\downarrow0.32}$&2.45$_{\uparrow1.98}$&0.47\\\cline {2-11}
&10. Mistreat&2.82$_{\uparrow0.69}$&3.93$_{\uparrow1.80}$&2.13&0.81$_{\downarrow1.84}$&2.49$_{\downarrow0.16}$&2.65&0.68$_{\downarrow0.35}$&2.34$_{\uparrow1.31}$&1.03\\ \cline {2-11}
 &11. Discrimination&7.18$_{\downarrow0.15}$&7.45$_{\uparrow0.12}$&7.33&5.73$_{\uparrow0.81}$&6.10$_{\uparrow1.18}$&4.92&6.59$_{\downarrow0.24}$&6.26$_{\downarrow0.57}$&6.83\\ \hline
 \end{tabular}}

\end{table}

Studies on bias in ML models have shown that bias is amplified in the model unless explicitly reducing bias-amplification \cite{zhao2017men,hall2022systematic}. Similarly, we see that the class distribution of the model predictions differs according to the label aggregation strategy used, with the effects of the data amplified in the model. Taking minority opinions into consideration can lead to over-predicting classes related to sexism. However, using minority aggregation and expert aggregation helps in classifying posts into the more subtle classes that are harder to identify, as well as those which constitute more harm to the receiver.

We suggest that rather than a single label aggregation strategy being ideal, it should be task-dependent, based on factors such as the objective of the task. For example, content moderation on a Twitch stream where no sexism is tolerated aims to capture all nuances of sexism, while content moderation policies on X or Facebook might be more lenient, where one might want to be more certain that a post is considered sexist by the majority of users. Considering whether a post might be viewed as sexist by the receiver depends on their personal sensitivities but may be more useful than considering the intent of the sender. 
\section{Minority Opinions Or Noise}
We conducted a qualitative analysis of the labels generated using minority aggregation. For this analysis, we have considered labels generated using minority aggregation as differing opinions of a minority of the annotators. However, as mentioned previously, some deviations from the majority or expert annotations may come from noisy annotations. We investigate whether the minority vote is noise or a valid point of view \cite{mokhberian2022noise,pavlick2019inherent}. We consider an annotation noisy if it is an arbitrary, spurious choice made due to misunderstanding the task, lack of focus, mistakes, etc., and having no logical reason or justification. We consider an annotation a valid point of view if the reasoning can be explained and justified.

We first collect all the posts from both EDOS and GE for which annotators disagree with each other. From these collections, we randomly sample 100 posts (i.e. 100 for EDOS, 100 for GE). We then categorise them as either differing opinions (valid) or noisy (noise) in two ways - manually and prompting an LLM. An expert assigned valid or noise labels to each post for manual analysis. We prompt GPT-3.5\footnote{\url{https://platform.openai.com/docs/models/gpt-3-5-turbo}} to generate valid and noise labels along with an explanation if a post is valid. The prompt we use is given below.

\textit{``Given the following text and sexism labels in the format text, label1, label2, label3. Is the minority-assigned sexism label valid? If yes, provide an explanation''}


There is 87.09\% alignment between manual labels and labels generated by GPT-3.5. 
Examples of valid disagreements and noise are in Table~\ref{tab:qual}.
\begin{table}[hbt!]
\caption{Examples of valid and noisy disagreements with explanations}
    \label{tab:qual}
    \resizebox{\textwidth}{!}{
    \centering
    \begin{tabular}
    {|p{1\linewidth}|p{0.2\linewidth}|p{0.2\linewidth}|p{0.2\linewidth}|}
    \hline
    \multicolumn{1}{|c}{Text} &\multicolumn{3}{|p{0.39\linewidth}|}{\centering Labels}\\\hline
    \textit{That evil woman deserves to be executed by hydrogen bomb.} \textbf{Valid}&none&none&2. Incitement\\
    Explanation: suggests extreme punishment, advocates use of hydrogen bomb to execute a woman&&&\\\hline
    \textit{You don't communicate being an alpha. You need to become an alpha. Women pick up on that far better than you can imagine.} \textbf{Valid}&none&none&7. Immutable diff\\
    Explanation: reinforces gender stereotypes by endorsing attractiveness of alpha masculinity. Oversimplifies male,female relationship preferences. Promotes a rigid hierarchy of male dominance that is harmful \& restrictive for both genders&&&\\\hline
    \textit{Thank God my wife is not like that, this lady is crazy} \textbf{Valid}&8. Backhand comp&none&none\\
    Explanation: speaker expresses relief that their wife lacks certain undesirable traits yet undermines another woman by calling her crazy, implying the superiority of the wife in a derogatory manner. This backhanded compliment arises from the contrast between gratitude towards the speaker's wife and the criticism of another woman.&&&\\\hline
    \textit{Men should go to war and be police officers and women should not. That is reality and I accept it. If that makes me a bad guy, so be it.} \textbf{Noise}&7. Immutable diff&8. Backhand comp&7. Immutable diff\\
    Explanation: no backhanded compliment in this statement&&&\\\hline
    \textit{Ugly women don't count} \textbf{Noise}&10. Mistreat&5. Dehumanise&none\\
    Explanation: perpetuates harmful stereotypes and contributes to a culture that objectifies and promotes discrimination and mistreatment based on superficial criteria. Therefore, the noisy label is ‘none’.&&&\\\hline
    \textit{fuck her off with a nine mm this is what you get when an arab fucks a monkey} \textbf{Valid and Noise}&4. Agg \& emo&none& 1. Threats\\
    Explanation: encourages harmful, aggressive sexual behaviour towards women and attacks both Arab people and women. In this case, we have a valid and noisy disagreement, making ‘none’ the noisy label’.&&&\\\hline
    \end{tabular}}
    
\end{table}

We find that for GE, 69\% of the disagreements are `valid', occurring for two main reasons. First, gender-based compliments (e.g. \textit{``women are beautiful like flowers. Delicate and deserving of tender loving care"}) may be considered positive or sexist depending on the annotator's context. Second, gendered slurs and profanity (such as `bitch' or `whore') used casually seem to cause disagreement. The reason for this is not always clear, though sometimes it may be due to misunderstood sarcasm (e.g. \textit{``I stay up till 2 smoke dope play video games, but she's the bitch for walking out on me"}). 

For EDOS, 76\% of the disagreements are `valid', mostly arising from annotators choosing one of many viable options. They frequently struggle to choose between derogation and 2 other classes - animosity and prejudice, and similarly between aggressive and emotive attacks (4) and slurs, profanities and insults (6). Again, sarcasm seems to be one of the main causes of noisy disagreements. In this dataset, we also found quoted tweets to be problematic since the comments were of the opposite point of view to the original message with which they were concatenated. For example, in the quoted tweet \textit{``women should be happy when men leer at them and catcall. It means they're attractive na na na YEAH SURE!"}, the capitalised comment at the end is disagreeing with the original message.

Following this analysis, we conclude that for EDOS and GE, most disagreements are indeed `valid', and thus, minority aggregation is generally a viable strategy for investigating the class distribution of minority voices in the data and comparing the label aggregation strategies. However, this depends on the quality of the dataset, and if the annotations are too noisy, we recommend first using a de-noising technique.


\section{Conclusion}
We have investigated the impact of 3 label aggregation strategies on the representation of minority opinions and the effect of using the gold labels generated by each on downstream model behaviour. We also verified via a qualitative study that most of the disagreements in our two sexism detection datasets arise from genuine differing opinions rather than noisy annotations, though this might not be the case in other datasets.

We conclude that each label aggregation strategy has its own bias and leads to different label distributions and model behaviour. Using majority aggregation will lead to a dataset and model that may exclude certain opinions, particularly those that are more sensitive to sexism. Using expert aggregation is more sensitive to sexism than majority aggregation (i.e. it will label more posts as sexist), but runs the risk of biasing the dataset towards a single expert`s opinion, so is also less inclusive. Incorporating minority opinions gives the most diverse and inclusive dataset but can lead the trained model to over-classify posts as sexist. 

Therefore, the label aggregation strategy should be chosen carefully, keeping in mind the objective of the task and use case. Moreover, one should be mindful of minority opinions when choosing a label aggregation strategy. Where feasible, we suggest performing at least an analysis of how the label distribution changes with the label aggregation strategy and comparing it with the minority aggregation to ensure that the chosen strategy does not introduce substantial biases against minority voices.

\section{Limitations}
Working with only 2 datasets limits the generalisability of our results, especially for datasets that contain many noisy annotations. In those cases, an additional noise removal step is required. Differing opinions and data ambiguity are general problems for all datasets, but fine-grained labels should be used cautiously. They are useful only if easily distinguishable (by humans as well as machines); otherwise, they unnecessarily contribute to disagreements which are less related to differing opinions and more about being forced to make a single choice, which can skew results to some extent. One should also be mindful about the choice of expert used in the case of expert arbitration for disagreements: in our study, only one dataset used an expert and therefore, the results of comparing an expert selection strategy with the other strategies might differ in other datasets.
\section{Ethics}
Our work is part of ongoing research into bias and fairness in sexism detection. While it deals with sensitive and offensive topics and language, it furthers the study of inclusivity and diversification of NLP to better serve and accommodate people from varying demographic groups. The datasets we use contain no sensitive or personal information about the annotators - any such information has already been either anonymised or not made publicly available by the original authors of the datasets. The work has undergone ethics approval from the authors' institution.
\begin{credits}
\subsubsection{\ackname} The study was conducted as part of the ``Responsible AI for Inclusive, Democratic Societies: A cross-disciplinary
approach to detecting and countering abusive language online'' project [grant number R/163157-11-
1]. 

\end{credits}

 
 
 

%
%
%
\bibliographystyle{splncs04}
\bibliography{custom.bib}
%


\appendix
\section{EDOS T1 and T2 Results}
\label{app:A}
\subsection{Investigating Aggregation on Data}
From Table~\ref{tab:ss_edos_model_app}, we observe that in T1, majority aggregation and expert aggregation are similar, also with fewer posts labelled sexist than minority aggregation. In T2, majority aggregation labels most posts as sexist for class derogation, while minority aggregation has the highest proportion for class prejudice. expert aggregation is highest for classes threats and animosity. 
\begin{table}
 \caption{Class distributions \% from test data \& model predictions \& difference from the original gold labels}
 \label{tab:ss_edos_model_app} 
\resizebox{\textwidth}{!}{
 \centering
 \begin{tabular}{|p{0.08\linewidth}|l|c|c|c|c|c|c|c|c|c|} \hline 
\multicolumn{1}{|c}{Dataset}& \multicolumn{1}{|c}{Class} &\multicolumn{3}{|c}{Data}& \multicolumn{3}{|c}{BERT} &\multicolumn{3}{|c|}{RoBERTa}\\ \hline
&& majority&minority&expert&majority&minority&expert&majority&minority&expert\\ \hline 
 \multirow{2}{0.08\linewidth}{\centering T1}&not sexist&75.62$_{\downarrow0.11}$&72.11$_{\downarrow3.62}$&75.73&77.36$_{\uparrow2.73}$&69.77$_{\downarrow4.86}$&74.63&80.54$_{\downarrow0.22}$&75.00$_{\downarrow5.76}$&80.76\\ \cline{2-11} 
 &sexist&24.38$_{\uparrow0.11}$&27.89$_{\uparrow3.62}$&24.27&22.64$_{\downarrow2.73}$&30.23$_{\uparrow4.86}$&25.37&19.46$_{\uparrow0.22}$&25.00$_{\uparrow5.76}$&19.24\\  \hline
 \multirow{4}{0.08\linewidth}{\centering T2}&threats&8.46$_{\downarrow0.76}$&8.12$_{\downarrow1.10}$&9.22&8.20$_{\downarrow0.28}$&5.45$_{\downarrow3.03}$&8.48&7.14$_{\downarrow0.34}$&4.32$_{\downarrow2.82}$&7.48\\ \cline{2-11}
&derogation&48.92$_{\uparrow4.81}$&44.20$_{\uparrow0.09}$&44.11&49.44$_{\uparrow1.12}$&49.74$_{\uparrow1.42}$&48.32&50.98$_{\uparrow1.55}$&51.11$_{\uparrow1.68}$&49.43\\ \cline{2-11}  &animosity&30.99$_{\downarrow4.11}$&33.88$_{\downarrow1.22}$&35.10&35.78$_{\uparrow1.6}$&34.11$_{\downarrow0.07}$&34.18&35.75$_{\downarrow0.47}$&35.03$_{\downarrow1.19}$&36.22\\ \cline{2-11} &prejudice&11.63$_{\uparrow0.05}$&13.80$_{\uparrow2.22}$&11.58&6.58$_{\uparrow0.56}$&10.70$_{\uparrow4.68}$&6.02&6.13$_{\downarrow0.74}$&9.54$_{\uparrow2.67}$&6.87\\\hline
\end{tabular}} 
\end{table}

\subsection{Investigating Aggregation on Models}
\begin{figure}[hbt!]
 \centering
 \subfigure{\includegraphics[width=0.4\linewidth,height=0.4\linewidth]{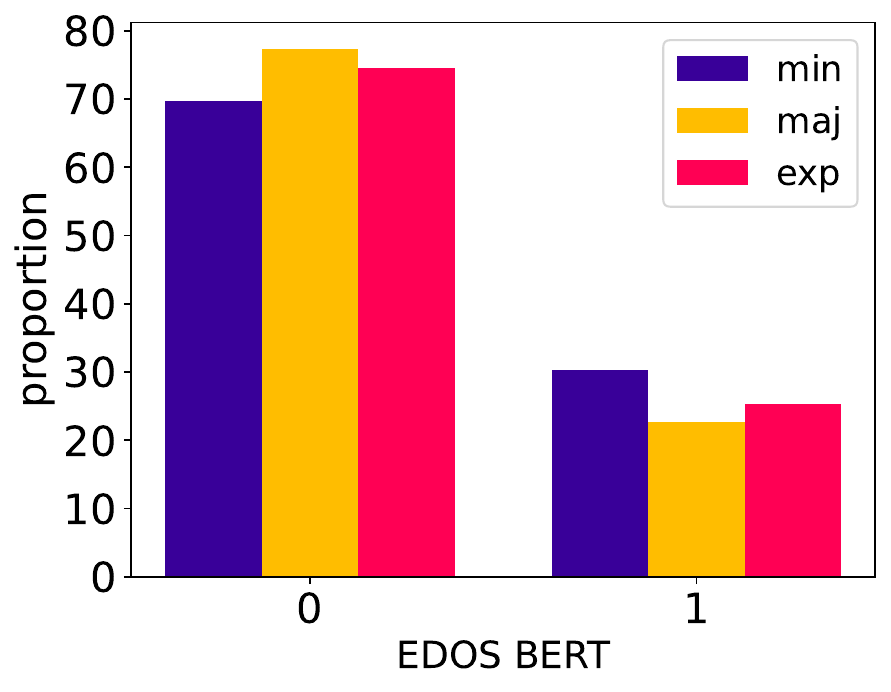}}
 \subfigure{\includegraphics[width=0.4\linewidth,height=0.4\linewidth]{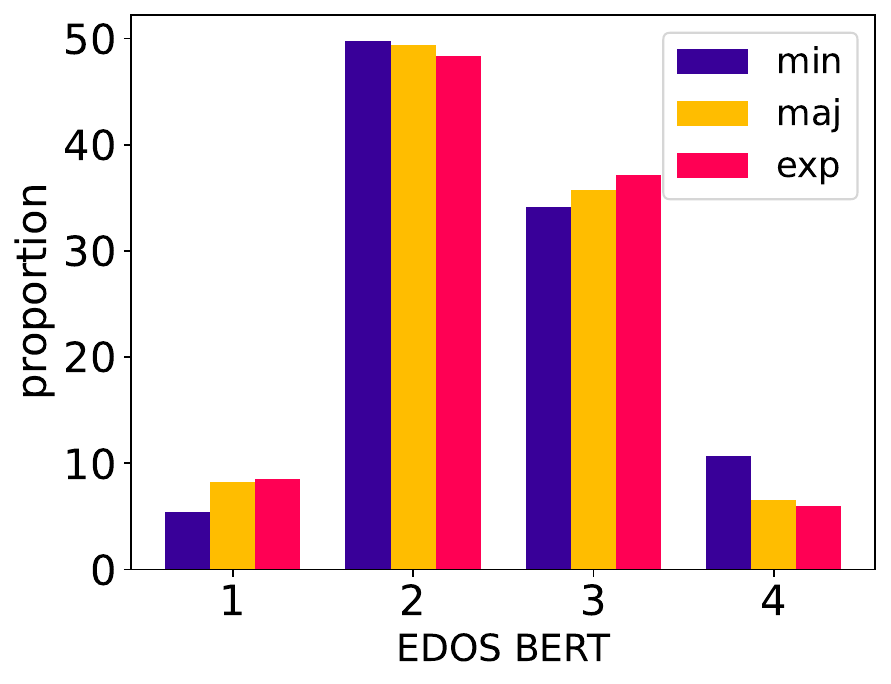}} 
 \caption{Class distributions of BERT predictions for EDOS T1 and T2, based on different label aggregation strategies}
 \label{fig:edos_ss_model_app}
\end{figure}

From Figure~\ref{fig:edos_ss_model_app} and Table~\ref{tab:ss_edos_model_app} in T1, majority aggregation has the lowest proportion of class sexist and minority aggregation the highest. In T2, majority aggregation and expert aggregation have a higher proportion of class threats and animosity than minority aggregation. Minority aggregation has a higher proportion of class prejudice.

\section{Model Details}
\label{app:D}
\subsection{Model Parameters}
For BERT, we use the base uncased model, and for RoBERTa, we use the base model. All models are trained on an NVIDIA A100 GPU. All models are run using 5-fold cross-validation with 4-fold data as training data and 1-fold data as testing data. The training data is split into train and validation sets (9:1). The maximum sequence length is 256 tokens, and the batch size is 32. We use Cross Entropy Loss as the training loss function with the AdamW optimizer. The learning rate of 5e-5. We choose the model with the smallest validation loss over 10 epochs during training.

\subsection{Performance Metrics}

For evaluation, we report average Precision, Recall and macro F1 over 5 folds with standard deviations. These can be found in Table~\ref{tab:f1}

\begin{table}[ht!]
 \caption{Performance Metrics}
 \label{tab:f1}
 \footnotesize
\resizebox{\textwidth}{!}{
 \centering
 \begin{tabular}{|p{0.08\linewidth}|l|c|c|c|c|c|c|c|c|} \hline 
 \multicolumn{10}{|c|}{BERT} \\ \hline 
\multicolumn{1}{|c}{Dataset}&\multicolumn{3}{|c}{Majority}& \multicolumn{3}{|c}{Minority} &\multicolumn{3}{|c|}{Expert}\\ \hline
&F1&Recall&Precision&F1&Recall&Precision&F1&Recall&Precision\\\hline 
GE&82.61$\pm$1.40&83.46$\pm$0.58&82.28$\pm$2.11&68.69$\pm$3.23&69.37$\pm$4.66&69.31$\pm$2.80&-&-&-\\\hline
T1&79.92$\pm$2.25&78.46$\pm$3.73&82.84$\pm$1.84&72.05$\pm$1.64&71.76$\pm$2.45&72.66$\pm$0.72&80.83$\pm$0.69&79.02$\pm$1.99&83.95$\pm$2.06\\\hline
T2&46.80$\pm$3.97&45.15$\pm$4.07&59.96$\pm$7.75&39.55$\pm$3.47&39.82$\pm$2.35&48.82$\pm$5.74&73.60$\pm$4.56&69.90$\pm$4.76&82.13$\pm$3.50\\\hline
T3&27.71$\pm$1.96&28.15$\pm$1.72&29.93$\pm$4.04&23.85$\pm$2.17&25.10$\pm$1.12&25.36$\pm$2.51&62.32$\pm$8.10&60.67$\pm$7.82&72.92$\pm$9.46\\\hline
\multicolumn{10}{|c|}{RoBERTa} \\ \hline 
\multicolumn{1}{|c}{Dataset}&\multicolumn{3}{|c}{Majority}& \multicolumn{3}{|c}{Minority} &\multicolumn{3}{|c|}{Expert}\\ \hline
&F1&Recall&Precision&F1&Recall&Precision&F1&Recall&Precision\\ \hline 
GE&83.98$\pm$2.17&85.52$\pm$1.33&83.26$\pm$2.26&67.70$\pm$4.62&69.19$\pm$5.17&69.37$\pm$1.82&-&-&-\\\hline
T1&80.12$\pm$3.25&78.65$\pm$2.18&81.71$\pm$2.48&71.80$\pm$3.33&70.39$\pm$1.42&73.14$\pm$1.29&78.63$\pm$1.89&78.02$\pm$0.79&81.87$\pm$2.10\\\hline
T2&47.51$\pm$3.74&46.72$\pm$3.20&55.17$\pm$2.92&37.19$\pm$2.89&39.62$\pm$2.09&44.47$\pm$5.53&66.27$\pm$8.57&65.47$\pm$7.70&72.49$\pm$9.35\\\hline
T3&26.16$\pm$1.42&27.93$\pm$2.64&29.42$\pm$3.87&22.38$\pm$1.29&24.36$\pm$0.54&22.18$\pm$2.25&63.29$\pm$5.08&58.31$\pm$7.71&69.95$\pm$6.78\\\hline
\end{tabular}} 
\end{table}

\end{document}